\documentclass[10pt,twocolumn,letterpaper]{article}

\usepackage{cvpr}
\usepackage{times}
\usepackage{epsfig}
\usepackage{graphicx}
\usepackage{amsmath}
\usepackage{amssymb}
\usepackage{multirow}

\cvprfinalcopy

\begin{document}

%%%%%%%%% TITLE
\title{ Temporal Extension of Scale Pyramid \\and Spatial Pyramid Matching for Action Recognition}

\author{Zhenzhong Lan, Xuanchong Li, Alexandar G. Hauptmann\\
Carnegie Mellon University\\
5000 Forbes Ave, Pittsburgh, PA 15213\\ 
{\tt\small lanzhzh, xcli, alex @cs.cmu.edu}
}

\maketitle
%\thispagestyle{empty}

%%%%%%%%% ABSTRACT
\begin{abstract}
Historically, researchers in the field have spent a great deal of effort to create image representations that have scale invariance and retain spatial location information. This paper proposes to encode equivalent temporal characteristics in video representations for action recognition. To achieve temporal scale invariance, we develop a method called temporal scale pyramid (TSP). To encode temporal information, we present and compare two methods called temporal extension descriptor (TED) and  temporal division pyramid (TDP) . Our purpose is to suggest solutions for matching complex actions that have large variation in velocity and appearance, which is missing from most current action representations. The experimental results on four benchmark datasets, UCF50, HMDB51, Hollywood2 and Olympic Sports, support our approach and significantly outperform state-of-the-art methods. Most noticeably, we achieve $65.0\%$ mean accuracy and $68.2\%$ mean average precision on the challenging HMDB51 and Hollywood2 datasets which constitutes an absolute improvement over the state-of-the-art by $7.8\%$ and $3.9 \%$, respectively.
\end{abstract}

%%%%%%%%% BODY TEXT
\section{Introduction}

\begin{figure*}
\centering
\begin{tabular}{cc}
\fbox{\includegraphics[height = 3.5cm, width=7.6cm]{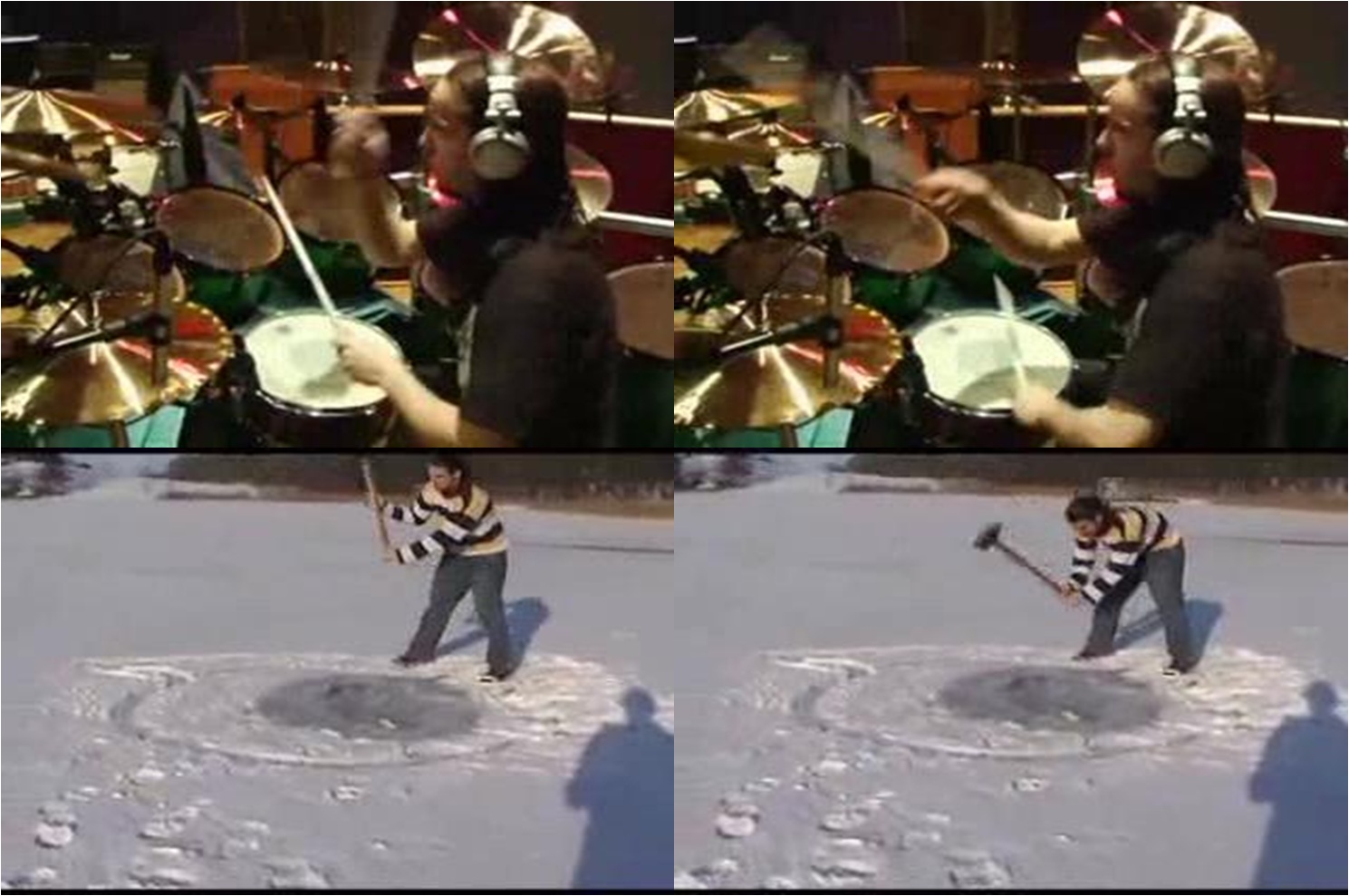}}&
\fbox{\includegraphics[height = 3.5cm,width=7.6cm]{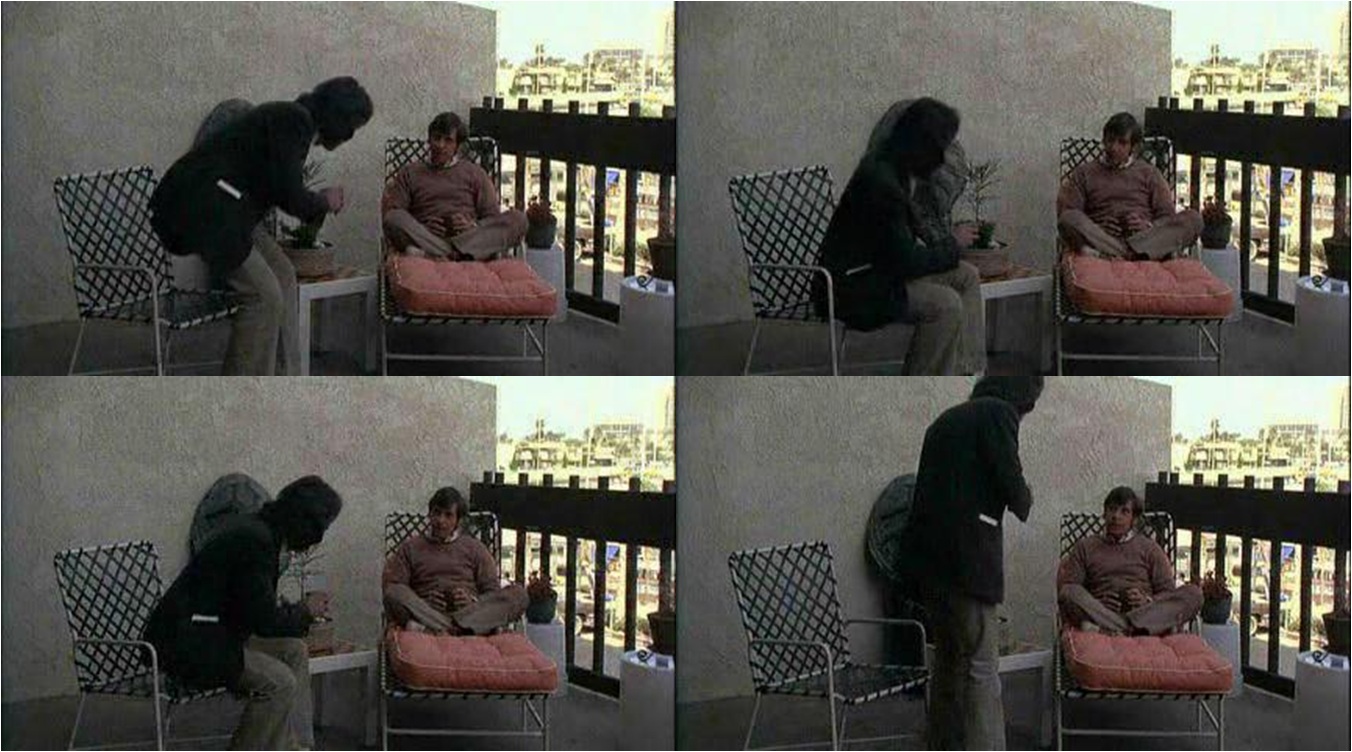}}\\
(a)&(b)
\end{tabular}
\caption{Example actions. (a) Playing drum (top) vs. digging a hole (bottom). (b) Sitting down (top) vs. standing up (bottom).}
\label{fig:illustration_action}
\end{figure*}

Consider the actions in Figure \ref{fig:illustration_action}. Shown in Figure \ref{fig:illustration_action} $(a)$ are two  ``hit'' actions that typically have different velocities. For playing the drum, people generally move the drum sticks quickly using  only their wrists. For digging a hole on a frozen river, people need to move their whole arms and upper body. Hacking the ice occurs at whatever pace the digger is comfortable with whereas drumming requires movement at precise intervals. Figure \ref{fig:illustration_action} $(b)$ shows two different actions, sitting down (top)  and standing up (bottom),  that have similar appearance. For most people, it is clear that the temporal order is important for discriminating between these actions. However, most current video representations remove temporal information. Due to the nature of actions and the way they are recorded, the illustrated phenomenon is ubiquitous. In this paper, we show how to take the velocity variation and temporal order into consideration to create better representations for complex human actions. 

Action recognition is becoming increasingly important in computer vision research due to its wide ranging applications which include human-computer interaction, health care and surveillance systems. Tremendous progress has been made in the accuracy of algorithms \cite{wang2011action, wang2013action, jain2013better,shi2013sampling} and the realism of datasets \cite{niebles2010modeling, kuehne2011hmdb,reddy2013recognizing,marszalek2009actions}. Nowadays, most action recognition datasets are real-world data rather than lab data \cite{blank2005actions,schuldt2004recognizing} . However, when performing recognition on these complex actions, where human pose estimation or template methods are not reliable, very few approaches explicitly take action velocity and temporal order into consideration. One approach to address velocity issue is to borrow from scale-space theory. 

\begin{figure*}
\centering
\begin{tabular}{ccc}
\fbox{\includegraphics[height = 2.5cm, width=5cm]{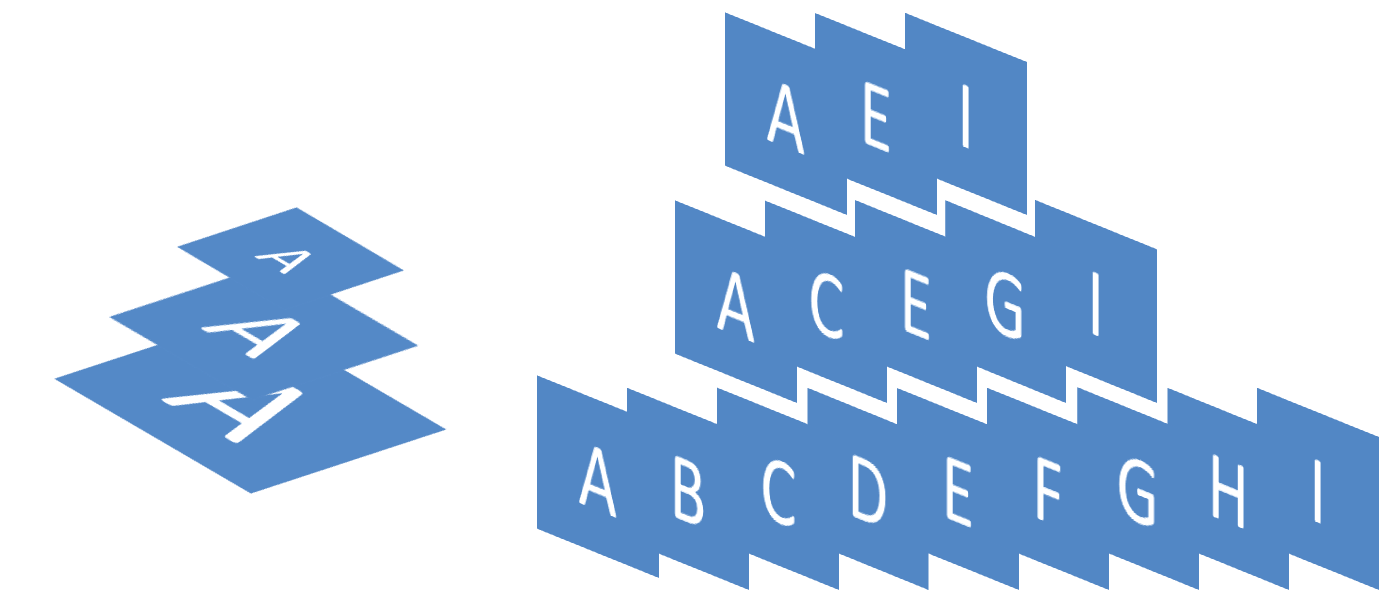}} 
&\fbox{\includegraphics[height = 2.5cm,width=5cm]{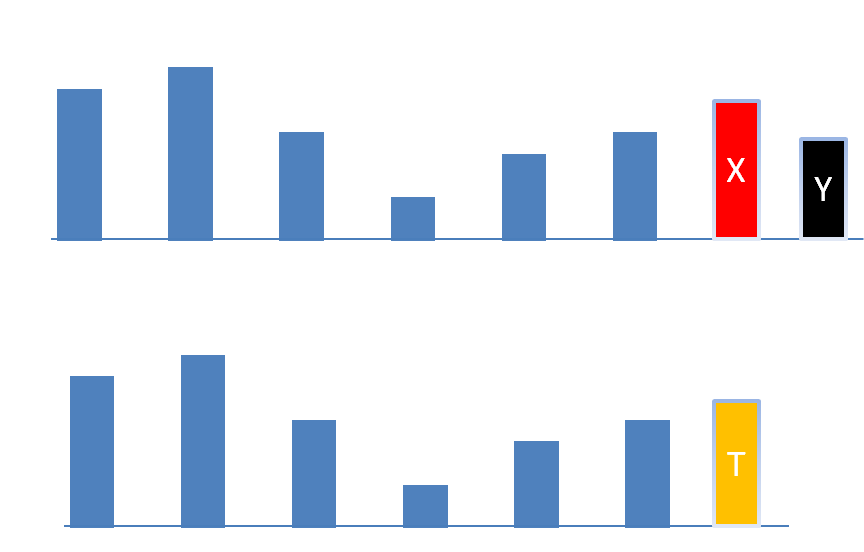}}&
\fbox{\includegraphics[height = 2.5cm,width=5cm]{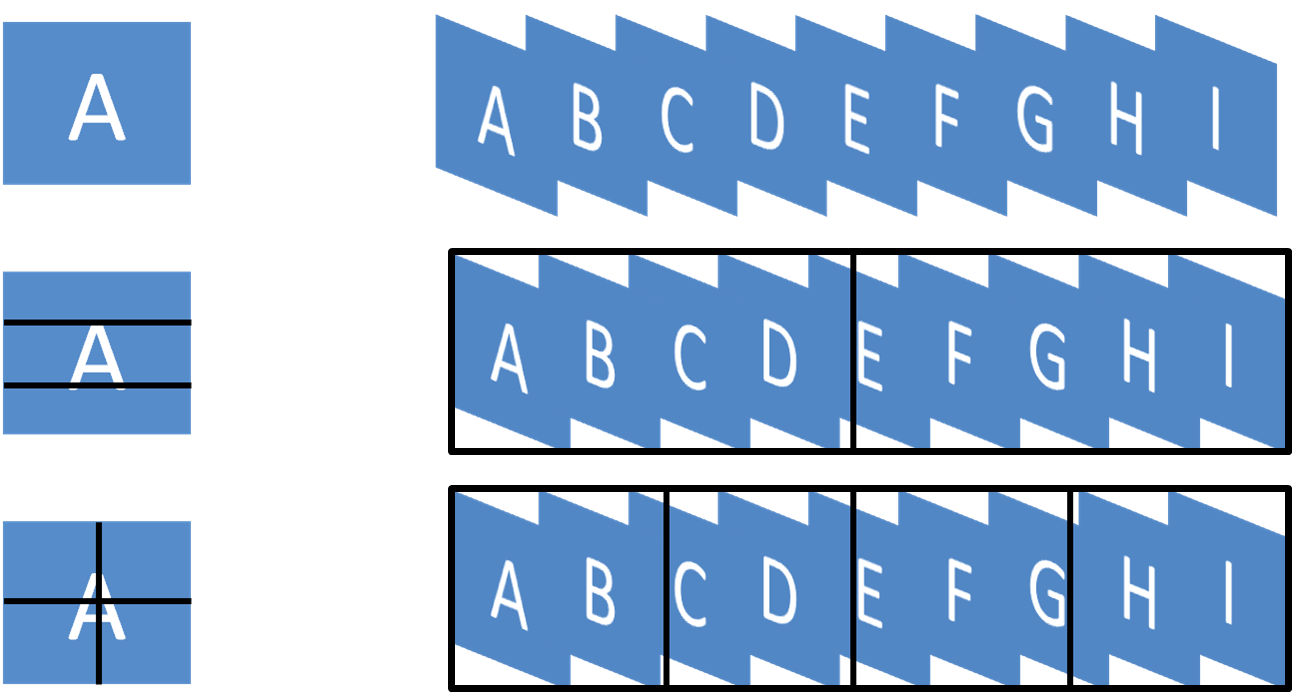}}\\
(a)&(b)&(c)
\end{tabular}
\caption{ (a) Image pyramid (left) vs. TSP (right); (b) spatial augmentation (top) vs. TED (bottom); (c) spatial pyramid (left) vs. TDP (right). }
\label{fig:illustration}
\end{figure*}

Scale-space theory \cite{lindeberg1994scale} states that it is crucial to use multi-scale representation for describing objects. We propose to develop an analogous temporal multi-scale representation for action recognition, as shown in Figure \ref{fig:illustration} (a), which we call \textit{temporal scale pyramid (TSP)}. Here we define the \textit{temporal scale} of an action as its \textit{velocity}. Temporal scaling allows us to match similar actions of different velocity. For example, a drummer drumming at 4 beats per second versus a digger moving at one stroke every 10 seconds. 

While actions have velocity, they also have ordered steps. For example, to perform an action called 'sitting down', we need to gradually bend our knees and lower our body, while 'standing up' is the reverse. Aggarwal and Ryoo \cite{aggarwal2011human} define these atomic actions as gestures. Inspired by the fact that actions can be segmented into consecutive gestures and different gesture orders may correspond to different actions, we develop two methods to encode temporal information in the original temporal insensitive video representations. For the first one, named \textit{temporal extension descriptor (TED)}, we add one dimensional temporal information into raw features, similar to spatial augmentation \cite{mccann2013spatially, sanchez2012modeling}, as shown in Figure \ref{fig:illustration} (b). The second one, which we call \textit{temporal division pyramid (TDP)},  mimics the spatial pyramid matching  \cite{lazebnik2006beyond} by dividing the videos into increasingly fine sub-temporal regions and computing statistics of local features found inside each sub-region, as shown in Figure \ref{fig:illustration} (c).  

\begin{figure*}
\centering
\begin{tabular}{c}
\includegraphics[height = 4cm, width=14cm]{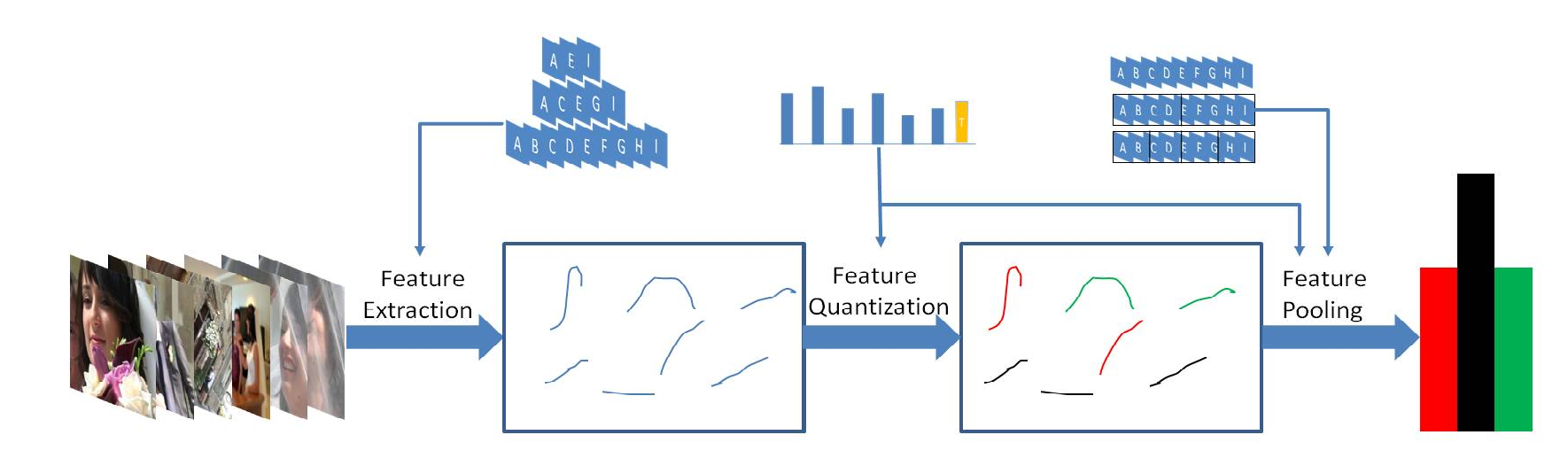} 
\end{tabular}
\caption{Feature extraction and representation pipeline and where our methods fit.}
\label{fig:pipeline}
\end{figure*}

Video representations using local features generally contain three stages, feature extraction, feature quantization and feature pooling.  TSP happens in the feature extraction phase, TED involves both feature quantization and feature pooling and TDP is a feature pooling technique, as shown in Figure \ref{fig:pipeline}. Because these three methods are applied at different stages in the pipeline, they are easy to combine. 

Our main contributions include proposing and examining the use of TSP, TED, TDP and combinations thereof and developing a working model for each case to achieve scale invariance and preserve temporal information for video representations. No complicated theory will be found in the following pages. All methods we propose are inspired by common sense and confirmed by our experiments on real-world datasets. 

\section{Related Work}
There is an extensive body of literature about action recognition, here we just mention a few revelant papers. See \cite{aggarwal2011human} for an in-depth survey. Features and encoding methods are two chief reasons for considerable progress in the field. Among them, the trajectory based approaches \cite{matikainen2009trajectons,sun2009hierarchical,wang2011action,wang2013action,jiang2012trajectory}, especially the Dense Trajectory method proposed by Wang et al. \cite{wang2011action,wang2013action}, together with the Fisher vector encoding  \cite{perronnin2010improving}  yields the current state-of-the-art performance on recognizing real-world human actions.

For lab datasets where human poses or action templates can be reliably estimated, dynamic time warping (DTP) \cite{ darrell1993space, veeraraghavan2006function}, hidden Markov models (HMMs) \cite{yamato1992recognizing, natarajan2007coupled} and dynamic Bayesian networks (DBNs) \cite{park2004hierarchical} are well studied methods for aligning actions that have velocity and motion appearance variation. However, for noisy real-world actions, these methods have not shown themselves to be very robust.

Pyramid methods \cite{adelson1984pyramid, lindeberg1994scale} have been very popular for most image processing tasks including image compression, image enhancement and object recognition. A multi-scale key-point detector proposed by \cite{lindeberg1993detecting} and used in SIFT \cite{lowe2004distinctive} to detect scale invariant key points using Laplacian pyramid method, in which Gaussian smoothing is used iteratively for each pyramid level. Dense SIFT \cite{vlfeat}  samples all sub-images from the original images using simple image resizing methods such as bilinear or bicubic interpolation without iterations. Shao et al. \cite{zhen2013spatio} also try to achieve scale invariance for action recognition, but they use 3-D Laplacian pyramid and extract global features using 3D Gabor filters, which only captures a single temporal scale for each video. 

There has also been a large amount of work in trying to build representations that keep spatial information of image patterns. Among them, spatial pyramid matching \cite{lazebnik2006beyond} is the most popular one. However, building spatial pyramids requires dimensions that are orders of magnitude higher than the original spatial invariant representations and hence make it less suitable for high dimensional coding methods such as Fisher vector \cite{perronnin2010improving} and VLAD \cite{arandjelovic2013all}. Spatial Fisher vector \cite{krapac2011modeling} and spatial augmentation \cite{mccann2013spatially, sanchez2012modeling} provide more compact representations to encode spatial information and show similar performance as spatial pyramid methods. Few approaches consider encoding global temporal information into video representations. Oneata et al. \cite{oneata2013action} show that better action recognition performance can be achieved by dividing videos into two parts and encoding each one separately. However, no further analysis about optimal divisions has been done in \cite{oneata2013action}. Codella et al. \cite{codella2012video} try to use temporal pyramid for event detection. They use $n$ temporal segments, where $n$ incrementally increases from 1 to 10. Within each temporal segment, they use max, min or average pooling to aggregate frame level features, which are very different from our motion features.

\section{Methods}

\subsection{Temporal Scale Pyramid (TSP)}

TSP consists of smoothing temporally and sub-sampling from different time intervals. For smoothing, we use a similar approach as \cite{vlfeat}, that is, for all levels, smooth temporally and sample directly from the original video. Using Gaussian smoothing, several smoothing scales were tested including $\alpha=0$ (no smoothing) and $\alpha=\infty$ (motion blur). However, we find that no smoothing always worked best. One possible reason is that our motion features require tracking and smoothing makes tracking harder. For sub-sampling, we test up to 5 scale levels, each level n corresponds to putting together the feature sets extracted from videos composed from every n+1 frames. More specifically, for each original smoothed video, we generated a set of $V$ videos, such that for each $v$ ranging from 0 to $V-1$, video $v$ is generated by taking every $v+1$ frames from the original video. For each video $v$, we can extract a set of local features $F_v$. For each level $V$, we take the union of the feature sets $F_{0,...,V}$ to represent the feature set for this pyramid level. For most of the videos tested, features cannot be extracted at levels higher than 5 because of the limited length of the videos or the difficulty of tracking with sparse frames. For longer videos with slower motions, higher level scaling may be beneficial. TSP allows us to match actions at different velocities.

\subsection{Temporal Extension Descriptor (TED)}
For TED, we augment the raw features by adding one dimensional normalized temporal information into each of the feature descriptors. The temporal information is based on the frame number where the feature occurs and normalized by the total number of frames in the video. For example, if each of the original feature points, $F$,  has a descriptor $D_1$ of dimension $d_1$ a descriptor $D_2$ of dimension $d_2$, then we will augment both $D_1$ and $D_2$ with the normalized $t$ that represents the relative temporal location of $F$ to form two new descriptors with dimensions $d_1+1$ and $d_2+1$, respectively. By using the TED strategy, we not only group temporally close feature points into the same cluster, but also enforce  action ordering on the video representation. 

\subsection{Temporal Division Pyramid (TDP)}
For TDP, given a video, we repeatedly divide the video into 2 separate temporal regions up to 8 sub-regions in total. We obtain the statistics of feature descriptors from each sub-region, and concatenate all statistics together to represent the video. For single-level comparison, we match solely on individual regions of that level. For the pyramid matching, we concatenate each of the single-level representations from the current and all lower levels. For example, let us assume that we want to construct a TDP for a video $v$ at level 2 using a bag-of-visual-word representation with $k$ centroids. We first map all the local feature descriptors extracted from $v$ into $k$ centroids to form a $k$ dimensional bag-of-visual-word representation $P_1$ for level 1. For level 2, we map the local feature descriptors from the first half of the video and the second half of the video into the same $k$ centroids to form another two bag-of-visual-word representations $P_{21}$ and $P_{22}$, respectively. For a single level-2 representation, we concatenate  $P_{21}$ and $P_{22}$ together to form a $2 \times k$ dimensional representation. For the pyramid level-2 representation, we concatenate $P_1$, $P_{21}$ and $P_{22}$ together to form a $3 \times k$ dimensional representation. TDP allows us to distinguish actions with different ordering.

\section{Experiments}

\subsection{Feature, Representation and Classification}

Improved Dense Trajectory with Fisher vector encoding \cite{wang2013action} represents a current state-of-the-art for most real-world action recognition datasets. Therefore, we use it to evaluate our methods. Note that although we use Improved Dense Trajectory and Fisher vector, our methods can be applied to any local features that involves optical flow calculation like STIP \cite{laptev2005space} and MoSIFT \cite{chen2009mosift} and any quantization and pooling methods such as VLAD \cite{arandjelovic2013all}. 

Our baseline method uses the  same settings as in \cite{wang2013action} except augmenting raw descriptors with spatial information as in \cite{mccann2013spatially}. These settings include the Improved Dense Trajectory feature extraction, Fisher vector representation and a linear SVM classifier.  

Improved Dense Trajectory features are extracted using 15 frame tracking, camera motion stabilization with human masking and RootSIFT normalization and described by Trajectory, HOG, HOF and MBH descriptors. We use PCA to reduce the dimensionality of these descriptors by a factor of two. After reduction, we augmented the descriptors with two dimensional normalized spatial location information as described in \cite{mccann2013spatially}. 

For Fisher vector representation,  we map the raw feature descriptors into a  Gaussian Mixture Model with 256 Gaussians trained from a set of randomly sampled 256000 data points. Power and L2 normalization are also used before concatenating different types of descriptors into a video based representation. 

For classification, we use a linear SVM classifier with a fixed C=100 as recommended by \cite{wang2013action} and the one-versus-all approach is used for multi-class classification scenario.

\subsection{Datasets}

\begin{figure*}
\centering
\begin{tabular}{cc}
\fbox{\includegraphics[height = 2cm, width=6.6cm]{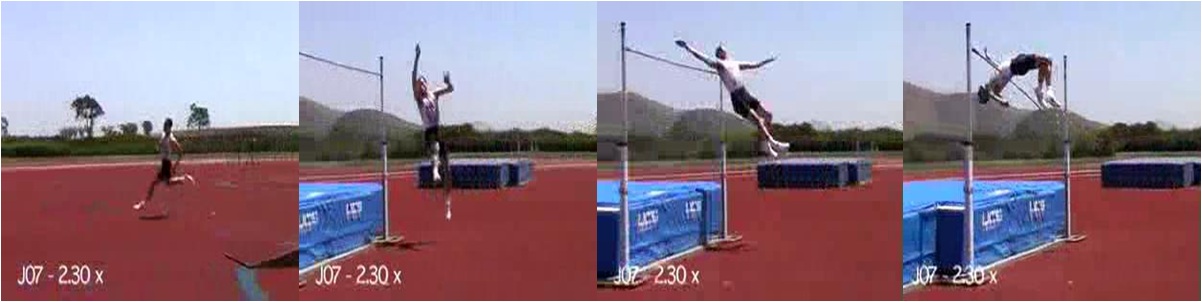}}&
\fbox{\includegraphics[height = 2cm,width=6.6cm]{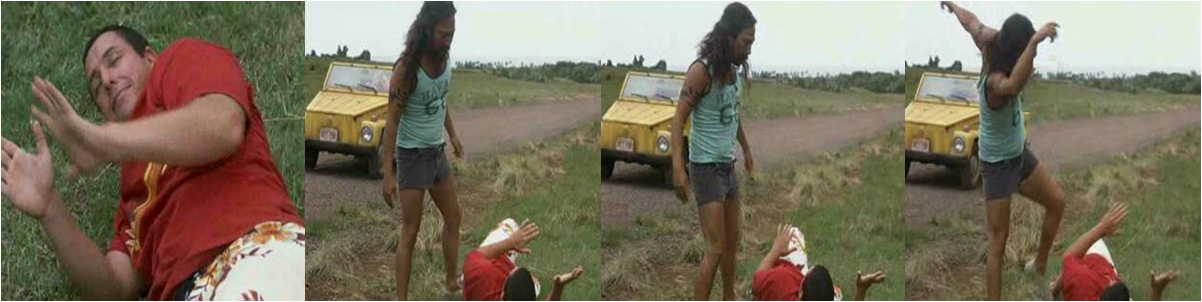}}\\
(a) HighJump &(b) Kick\\
\fbox{\includegraphics[height = 2cm, width=6.6cm]{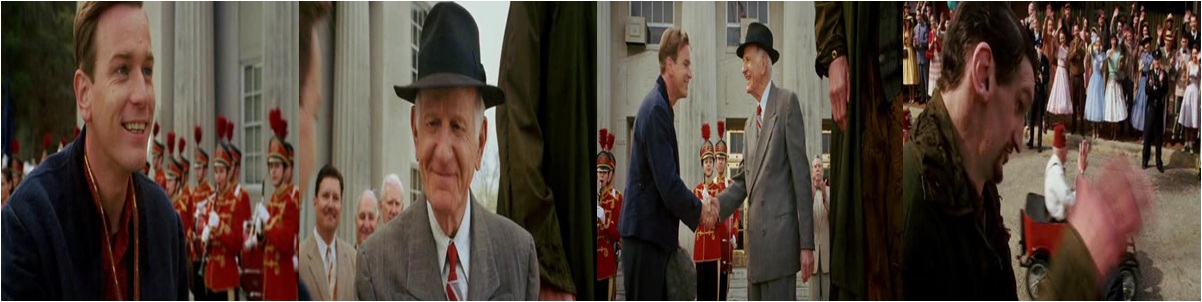}}&
\fbox{\includegraphics[height = 2cm,width=6.6cm]{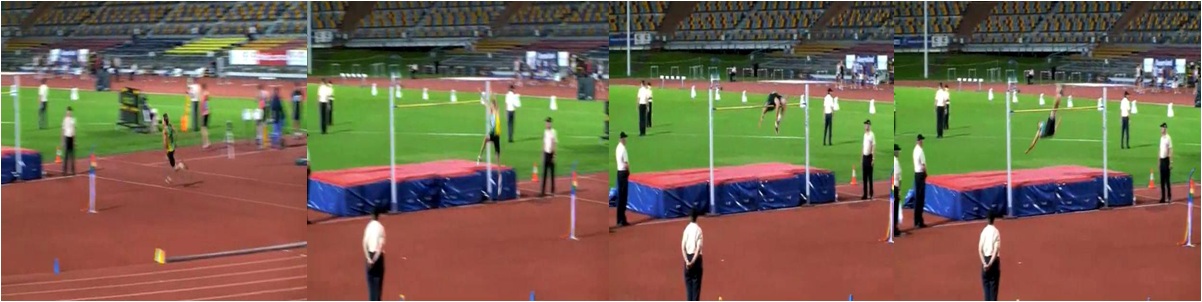}}\\
(c) HandShake&(d)HighJump
\end{tabular}
\caption{Examples from (a) UCF50, (b) HMDB51, (c) Hollywood2, (d) Olympic Sports.}
\label{fig:examples}
\end{figure*}

We use four action recognition datasets, UCF50, HMDB51, Hollywood2 and Olympic Sports, for evaluation. 
These datasets are selected because they are the real-world action datasets that have received the bulk of experimental attention, and because they reveal important aspects of our proposed methods.

The UCF50 dataset \cite{reddy2013recognizing} has 50 action classes spanning over 6618 YouTube videos clips that can be split into 25 groups. The video clips in the same group are generally very similar in background. Leave-one-group-out cross-validation as recommended by \cite{reddy2013recognizing} is used and mean accuracy (mAcc) over all classes and all groups is reported.

The HMDB51 dataset \cite{kuehne2011hmdb} has 51 action classes and 6766 video clips extracted from digitized movies and YouTube. \cite{kuehne2011hmdb} provides both original videos and stabilized ones. We only use original videos in this paper and standard splits with mAcc are used to evaluate the performance.

The Hollywood2 dataset \cite{marszalek2009actions} contains 12 action classes and 1707 video clips that are collected from 69 different Hollywood movies. We use the standard splits with training and test videos provided by \cite{marszalek2009actions}.   Mean average precision (mAP) is used to evaluate this dataset because multiple labels can be assigned to one video clip.

The Olympic Sports dataset \cite{niebles2010modeling} consists of 16 athletes
practicing sports, represented by a total of 783 video clips. We use standard splits with 649 training clips and 134 test clips and report mAP as in \cite{niebles2010modeling} for comparison purposes. 

In Figure \ref{fig:examples}, we show some example frames from the four datasets. Figure \ref{fig:examples} (a) and (d) from UCF50 and Olympic Sports show examples of continuity of perspective while Figure \ref{fig:examples} (b) and (c) from HMDB51 and Hollywood2 show how perspectives can change when the data comes from movies. 

\begin{table}
\centering
\begin{tabular}{|c |c |c |c|  }
\hline
 Datasets  & Source& Durations & mTSVF \\\hline
UCF50 &YouTube & 7.44s & 0.21 \\
HMDB51  &YouTube/Movie & 3.14s & 0.43 \\
Hollywood2  &Movie& 11.55s & 0.60 \\
Olympic  &YouTube & 7.74s & 0.45 \\
\hline
\end{tabular}
\caption{\label{tab:dataset}Meta data for action datasets.}
\vspace{-5mm}
\end{table}

Since datasets are an integral part of action recognition research, not just as a source of comparing our algorithms to state-of-the-art approaches, but also as a way of understanding our methods, we would like to compare the different datasets we used and `predict' how well our algorithms will perform on these datasets. However, there are many factors about a dataset that can influence an algorithm's performance. For example, what kinds of actions are in a dataset, or how much training data is there. Here, from a very high level, we would predict our methods' relative performance improvement on the different datasets based on their meta data as shown in Table \ref{tab:dataset}, in which we show the source of the clips, average duration in seconds and mean temporal scale variation factor (mTSVF). The mTSVF is measured by the average of the within-class duration's standard deviation divided by its mean. A TSVF for an action class is propotion to the scale number needed to cover the velocity range for that class and we use it to measure the relative scale need for that action. We use a video clip's duration to approximate the inverse velocity of the action in that video clip. Although this is a rough approximation, especially for long duration clips that involve multiple actions or repeated actions, our hope is that by measuring on multiple noisy but large videos, we would be able to deduce some meaningful statistics of the datasets. Besides scale variation, data source is another important factor that can affect our algorithm's performance. Generally speaking, videos from movies are harder to recognize than videos from YouTube, because movies generally combine shots from multiple cameras and involve a lot of perspective changes and because movies generally have more complex scenes, whose features can dominate the target action features. 

Since our TSP method is designed for handling temporal scale variation, we expect that it will help more for those datasets that have higher scale variation. Thus, we predict that TSP will have higher performance improvement for HMDB51 than for UCF50 and Hollywood2 will be higher than Olympic Sports. Similarly, since TED and TDP are designed for helping to discriminate actions that have similar global appearance and different local appearance, we expect that it will help more on those datasets that have more complex scenes, where background noise can dominate global appearances. Because HMDB51 and Hollywood2 have more complex scenes than UCF50 and Olympic Sports, we again predict that the improvement of TED and TDP methods for HMDB51 and Hollywood2 will be higher than for UCF50 and Olympic Sports, respectively. 
 
\subsection{Experimental Results}
\subsubsection{Temporal Scale Pyramid (TSP)}

\begin{table}
\centering
\begin{tabular}{|c | c |c |c |c|}
\hline
  & UCF50  & HMDB51 & Hollywood2 & Olympic \\
  Level& (mAcc.$\%$) & (mAcc.$\%$) & (mAP$\%$) & (mAP$\%$)\\\hline 

 0  & 92.8 &  59.4 & 65.6 & 89.6\\
 1 & 93.7 & 63.8 & 66.1 & \textbf{91.9}\\
 2 & \textbf{93.9}  &  64.0 & 66.5 & \textbf{91.9} \\
 3 & 93.8  & 64.3 & \textbf{66.6} & 90.5\\
 4 & 93.8  &  \textbf{64.4} & 66.0 & 90.5 \\
 5 & 93.7 & \textbf{64.4} & 65.8 & 89.9\\
\hline
\end{tabular}
\caption{\label{tab:scale Pyramid}Comparison of different scale levels for TSP.}
\end{table}

We evaluate the impact of the scale level on the TSP strategy with results in Table \ref{tab:scale Pyramid}. We compare the performance of TSP at different scale levels. Level 0 corresponds to the original video. From Table \ref{tab:scale Pyramid}, we can see that compared with level 0, the biggest performance increase comes from level 1 pyramids. After that, the performance improvement becomes less prominent. For some cases such as Hollywood2 and Olympic Sports, at the higher levels, performance even decreases. This decrease may be because Improved Dense Trajectory feature requires tracking and if the frames are too sparse, tracking becomes unreliable and noisy trajectories can be introduced. Additionally, we observe that the performance peaks of HMDB51 and Hollywood2 are later than UCF50 and Olympic Sports, respectively, this may indicate that both HMDB51 and Hollywood2 require higher levels of scale due to their higher scale variation. We can also see that, as we expected, the improvement on HMDB51 is larger than the improvement on UCF50. However, the improvement on Hollywood2 is smaller than the improvement on Olympics Sports, which differs from our expectation. This difference may be because these two datasets are not comparable due to the fact that videos in Hollywood2 dataset can contain multiple actions. To remove this factor and further test if the TSP does achieve temporal scale invariance, we calculate the mTSVF for those action classes that have improvement or stay the same and those actions that have negative improvement and show the results in Table \ref{tab:scale Pyramid variance}.  From Table \ref{tab:scale Pyramid variance}, we can see that indeed those actions where TSP improves the performance are in  classes that have greater temporal scale range (large TSVF). Also, in Table \ref{tab:scale Pyramid variance}, we show that the number of classes where TSP strategy helps ($\Delta P \geq 0$) is significantly greater than those that our TSP strategy hurts ($\Delta P < 0$).

\begin{table}
\begin{tabular}{|c | c |c |c |c|}
\hline
& \multicolumn{2}{|c|}{\# classes} & \multicolumn{2}{|c|}{mTSVF}\\
 \hline
 & $\Delta P \geq 0$  & $\Delta P < 0$ & $\Delta P \geq 0$&$\Delta P < 0$ \\
 \hline
 UCF50 & 40 & 10 & 0.28 & 0.21 \\
 \hline
 HMDB51 & 46 & 5 & 0.55 & 0.41 \\
 \hline 
 Hollywood2 & 9 & 3 & 0.64 & 0.47 \\
 \hline 
 Olympic & 13 & 3 & 0.46 & 0.44 \\
 \hline
\end{tabular}

\caption{\label{tab:scale Pyramid variance}Number of classes that have improved \textbf{P}erformance ($\Delta P \geq 0$), and have negative improvement ($\Delta P < 0$) using the TSP strategy and the mTSVF of  those classes that have improvement ($\Delta P \geq 0$) and have negative improvement ($\Delta P < 0$).} 
\end{table}

\subsubsection{Temporal Extension Descriptor (TED)}
For TED, we simply add one dimensional normalized temporal information into each raw Dense Trajectory descriptor. From  Table \ref{tab:temporal augmentation}, we see that this addition improves the accuracy for all datasets. However, the most significant improvement occurs with the complex datasets. This improvement may indicate that, as we expected, our temporal encoding methods are more suitable for action datasets that involve complex scenes.

\subsubsection{Temporal Division Pyramid (TDP)}
In Table \ref{tab:temporal pyramids}, we compare the performance of the TDP at different levels. Level 1 corresponds to the original video representations without division. We test up to 8 divisions of the videos. From Table  \ref{tab:temporal pyramids}, we can see that the TDP strategy only helps for the videos that have a large number of perspective changes. Further division (from level 2 to 4 or 8) almost always results in worse performance with the exception of the pyramid in the HMDB51 dataset. These results show that encoding temporal information does help to improve the performance yet it is very hard to find a balance between temporal translation invariance and temporal information encoding. Comparing Tables \ref{tab:temporal pyramids} and \ref{tab:temporal augmentation}, we can see that TED is a better temporal information encoding method in our setting. 

\begin{table}
\centering
\begin{tabular}{|c| c|c |c |c|}
\hline
  & UCF50  & HMDB51 & Hollywood2 & Olympic \\
 TED & (mAcc. $\%$)& (mAcc. $\%$) & (mAP $\%$) & (mAP $\%$)\\ \hline
 w/o &92.8  & 59.4 & 65.6 & 89.6  \\
 w & \textbf{93.0} &  \textbf{62.1} & \textbf{67.0} & \textbf{89.8} \\
 \hline
\end{tabular}
\caption{\label{tab:temporal augmentation}Performance of TED.}
\end{table}

\begin{table*}
\centering
\begin{tabular}{|c| c| c |c |c | c|c |c |c|}
\hline
  & \multicolumn{2}{|c|}{UCF50}  & \multicolumn{2}{|c|}{HMDB51} & \multicolumn{2}{|c|}{Hollywood2} & \multicolumn{2}{|c|}{Olympic} \\
  & \multicolumn{2}{|c|}{(mAcc. $\%$)} & \multicolumn{2}{|c|}{(mAcc. $\%$)} & \multicolumn{2}{|c|}{(mAP $\%$)} & \multicolumn{2}{|c|}{(mAP $\%$)}\\ \hline
  Level & Single& Pyramid & Single & Pyramid & Single& Pyramid &Single& Pyramid\\
  \hline
 1 &\textbf{92.8} &  & 59.4& & 65.6 & & \textbf{89.6} & \\
 2 &92.3& 92.6  & 61.0& 61.3 & 66.3 & \textbf{67.4} & 87.8 & 89.3\\
 4 &91.6  & 92.3 & 60.5& \textbf{61.6} & 65.2 & 67.0 & 85.2 & 87.6\\
 8 &90.4  & 92.1 & 58.1& \textbf{61.6} & 62.2 & 65.7 & 83.8 & 83.8\\
\hline
\end{tabular}
\caption{\label{tab:temporal pyramids}Comparison of different pyramid levels for TDP.}
\end{table*}

\subsubsection{Comparing with the State-of-the-Art}
In this section, we combine the three proposed methods and give the results in Table \ref{tab:state-of-art}, where we also compare these combined approaches with published state-of-the-art approaches. From Tables \ref{tab:scale Pyramid} and \ref{tab:temporal pyramids}, we see that scale level of 2 gives stable results for both TSP and TDP, therefore we set the scale level as 2 for both methods. From Table \ref{tab:state-of-art}, in all of the datasets, we see substantial improvement over the state-of-the-art. Especially on the two most challenging sets we improve the state-of-the-art performance of HMDB51 by $7.8\%$ and Hollywood2 by $3.9\%$.  Note that although we list several most recent approaches here for comparison purposes, most them use different features hence they are not directly comparable to our results. Shao et al. \cite{zhen2013spatio} propose to use Laplacian pyramid to encode temporal scale and get $37.3 \%$ accuracy on the HMDB51 dataset. Shi et al. \cite{shi2013sampling} use random sampled feature points and HOG, HOF, HOG3D and MBH descriptors. Jain et al. \cite{jain2013better}'s approach incorporates  a new motion descriptor. Oneata et al.  \cite{oneata2013action} focus more on testing Spatial Fisher vector for multiple action and event tasks. The most comparable one is Wang et al. \cite{wang2013action}, from which we build our approaches and which serves as our baseline. Of the combined methods, TSP + TED gives the best results for most of the datasets. TSP + TED + TDP gives better performance for the Hollywood2 dataset, however, combing TDP with any other method significantly increases the size of feature vector. This suggests that TSP + TED is a better combination to use in general.

\begin{table*}
\centering
\begin{tabular}{|c|c |c |c |c|}
\hline
  & UCF50  & HMDB51 & Hollywood2 & Olympic \\
  & (mAcc. $\%$) & (mAcc. $\%$) & (mAP $\%$) & (mAP $\%$)\\\hline 
  Shao et al. 13 \cite{zhen2013spatio} & N/A& 37.3 & N/A & N/A \\
Shi et al. 13 \cite{shi2013sampling} & 83.3& 47.6 & N/A & N/A\\
Jain et al. 13 \cite{jain2013better} & N/A &  52.1& 62.5 & 83.2 \\
Oneata et al. 13 \cite{oneata2013action} &90.0 &54.8 & 63.3 & 89.0 \\
Wang et al. 13 \cite{wang2013action} & 91.2 & 57.2 & 64.3 & 91.1\\
\hline 
TSP + TED & \textbf{94.2} & \textbf{65.0} &  67.9 & \textbf{92.9}\\ 
TSP + TDP & 93.8 & 64.7 &  68.0 & 92.3\\ 
TED + TDP & 93.0 & 61.8 &  67.1 & 89.9 \\ 
TSP + TED + TDP & 94.0 & 64.8 &  \textbf{68.2} & 92.5\\ \hline
\end{tabular}
\caption{\label{tab:state-of-art}Comparison of our results to the state-of-the-art.}
\end{table*}

\subsection{Computational Complexity}

The computational complexity of TED and TDP is negligible. The most significant computational cost is for TSP. Level 0 of TSP has the same cost as other single pass methods, for example, Wang et al. \cite{wang2013action}.  For level $l$, the cost becomes $1/l$ of the level 0. So with a TSP up to level 2, the computational cost will be less than twice the cost of a single pass through the original video, yet it can yield a significant improvement over current state-of-the-art methods.

\section{Discussion}

The history of video processing has witnessed numerous success stories of borrowing insight from the still image processing domain and applying it to the video domain.  We try to do similar things in this paper. Inspired by the success of image pyramids, spatial augmentation and spatial pyramids for object recognition, we propose to use TSP, TED, TDP to achieve temporal scale invariance and encode temporal information for video representations. For each case, we develop a corresponding working model and test it on four widely used action recognition datasets. Are these models successful? In some ways, yes. A combination of these models improves four benchmark datasets from the state-of-the-art significantly. According to our experiments, for general action recognition usage, we see great benefit in using TSP with a temporal scale level of 2 together with TED. Yet, these models do not have rigid mathematical derivations and their details are subject to change in light of new evidence, especially for temporal information encoding. We expect that greater improvement can be achieved if a better way of balancing temporal translation invariance and temporal information encoding can be found. Finally, if better local features, quantization or pooling methods are found in the future, our methods can still be used with them. 

{\small
\bibliographystyle{ieee}
\bibliography{egbib}
}
\end{document}